\documentclass[letterpaper, 10 pt, journal, twoside]{ieeetran}
\usepackage{cite}

\IEEEoverridecommandlockouts                              

\usepackage{graphicx}
\usepackage{setspace}
\usepackage{acronym}
\usepackage{amsfonts}			
\usepackage{times}		
\usepackage{color,soul}
\usepackage{amssymb}
\usepackage{amsmath}
\usepackage{gensymb}
\usepackage{tabularx} 
\usepackage{multirow}
\usepackage{algorithm}
\usepackage{algorithmic}
\usepackage{booktabs}
\usepackage{physics}
\usepackage[T1]{fontenc}

\usepackage{float}
\usepackage{mwe}
\newcolumntype{L}{>{\centering\arraybackslash}m{3cm}}
\usepackage[thinc]{esdiff}
\usepackage[table]{xcolor}
\usepackage{textcomp}
\usepackage{booktabs}
\usepackage{hyperref}
\hypersetup{
    colorlinks=true,
    citecolor=black,
    linkcolor=black,
    filecolor=black,      
    urlcolor=black,
    }
\usepackage{url}
\usepackage{array,ragged2e}
\newcolumntype{P}[1]{>{\RaggedRight\arraybackslash}p{#1}}

\newcolumntype{M}[1]{>{\centering\arraybackslash}m{#1}}

\usepackage[font=small,skip=2pt]{caption}
\usepackage{textcomp, gensymb}
\usepackage{dblfloatfix}
\usepackage{microtype}
\renewcommand{\baselinestretch}{0.955}
\usepackage{etoolbox}
\newbool{showcolor}
\setbool{showcolor}{true}
\newcommand{\changed}[1]{%
  \ifbool{showcolor}{\textcolor{blue}{#1}}{#1}%
}

\begin{document}

\title{
SCANS: A Soft Gripper with Curvature and Spectroscopy Sensors for \\In-Hand Material Differentiation
}

\author{Nathaniel Hanson$^{1,2*}$, Austin Allison$^{1*}$, Charles DiMarzio$^{1}$, Taşkın Padır$^{1,3}$, Kristen L. Dorsey$^{1}$
\thanks{Manuscript received: April 11, 2025; Revised August 8, 2025; Accepted September 24, 2025.}
\thanks{This paper was recommended for publication by Editor Cecilia Laschi upon evaluation of the Associate Editor and Reviewers' comments.}
\thanks{This research is supported by NSF Award Number 1928654.}
\thanks{$^{1}$Nathaniel Hanson, Austin Allison, Charles DiMarzio, Taşkın Padır, and Kristen L. Dorsey are with the Electrical and Computer Engineering Department, Northeastern University, Boston, Massachusetts, USA.}
\thanks{$^{2}$Nathaniel Hanson is with the Lincoln Laboratory, Massachusetts Institute of Technology, Lexington, Massachusetts, USA.}
\thanks{$^{*}$Equal Contribution; Correspondence: {\tt\footnotesize nhanson2@mit.edu}.}%
\thanks{$^{3}$Ta\c{s}k{\i}n Pad{\i}r holds concurrent appointments as a Professor of Electrical and Computer Engineering at Northeastern University and as an Amazon Scholar. This paper describes work performed at Northeastern University and is not associated with Amazon.}
}
\markboth{IEEE Robotics and Automation Letters. Preprint Version. Accepted September, 2025}
{Hanson \MakeLowercase{\textit{et al.}}:  A Soft Gripper with Curvature and Spectroscopy Sensors for In-Hand Material Differentiation}

\maketitle
\begin{abstract}
We introduce the soft curvature and spectroscopy (SCANS) system: a versatile, electronics-free, fluidically actuated soft manipulator capable of assessing the spectral properties of objects either in hand or through pre-touch caging. This platform offers a wider spectral sensing capability than previous soft robotic counterparts. We perform a material analysis to explore optimal soft substrates for spectral sensing, and evaluate both pre-touch and in-hand performance. Experiments demonstrate explainable, statistical separation across diverse object classes and sizes (metal, wood, plastic, organic, paper, foam), with large spectral angle differences between items. Through linear discriminant analysis, we show that sensitivity in the near-infrared wavelengths is critical to distinguishing visually similar objects. These capabilities advance the potential of optics as a multi-functional sensory modality for soft robots. The complete parts list, assembly guidelines, and processing code for the SCANS gripper are accessible at: \url{https://parses-lab.github.io/scans/}.

\end{abstract}

\begin{IEEEkeywords}
Soft Sensors and Actuators; Soft Robot Applications; Grippers and Other End-Effectors
\end{IEEEkeywords}

\section{Introduction}

\IEEEPARstart{S}{oft} robot grippers are increasingly used in commercial, healthcare, and industrial applications for material handling. While commercial and research-grade sensors have improved force distribution and proprioception, they still lack the ability to distinguish subtle differences in material properties---texture, moisture content, or internal structure---between visually similar objects. Moreover, soft grippers often lack sensing modalities that allow them to perceive object properties before contact. This absence of pre-touch sensing limits their ability to anticipate how best to grasp, manipulate, or sort objects, especially when visual information is ambiguous or occluded. Access to material composition before or during grasping would allow soft grippers to more accurately classify objects and infer functional properties such as ripeness or fragility, thereby enhancing manipulation, planning, and execution.

Spectral analysis of material composition, such as in the near-infrared (NIR) and short-wave infrared spectrum at wavelengths from $800-2500$ nm \cite{pasquini2018near}, is a widely used set of techniques in laboratory environments. Spectrometers measure the intensity of the reflected light spectrum created by light-surface interactions.
When compared to the illumination source, the patterns of peaks or troughs in the spectrum reveal the presence of particular chemical bonds and therefore the composition of the material (e.g., water or sugar content) \cite{hanson2022slurp}.

In traditional robotics, material recognition has been explored using visual and non-visual methods \cite{erickson2020multimodal}, yet existing techniques rely on rigid sensors that are not compatible with soft grippers. 
Sensors fabricated from compliant materials \cite{hegde2023sensing} largely focus on measuring mechanical forces,
but distinguishing chemically or visually similar materials remains a challenge that must be solved to translate such approaches to practical applications.
These gaps underscore a fundamental need in robotics: \textit{mechanically robust and compliant sensors that perceive material properties.} 

\begin{figure}[!tbp]
  \centering
  \includegraphics[width=\linewidth]{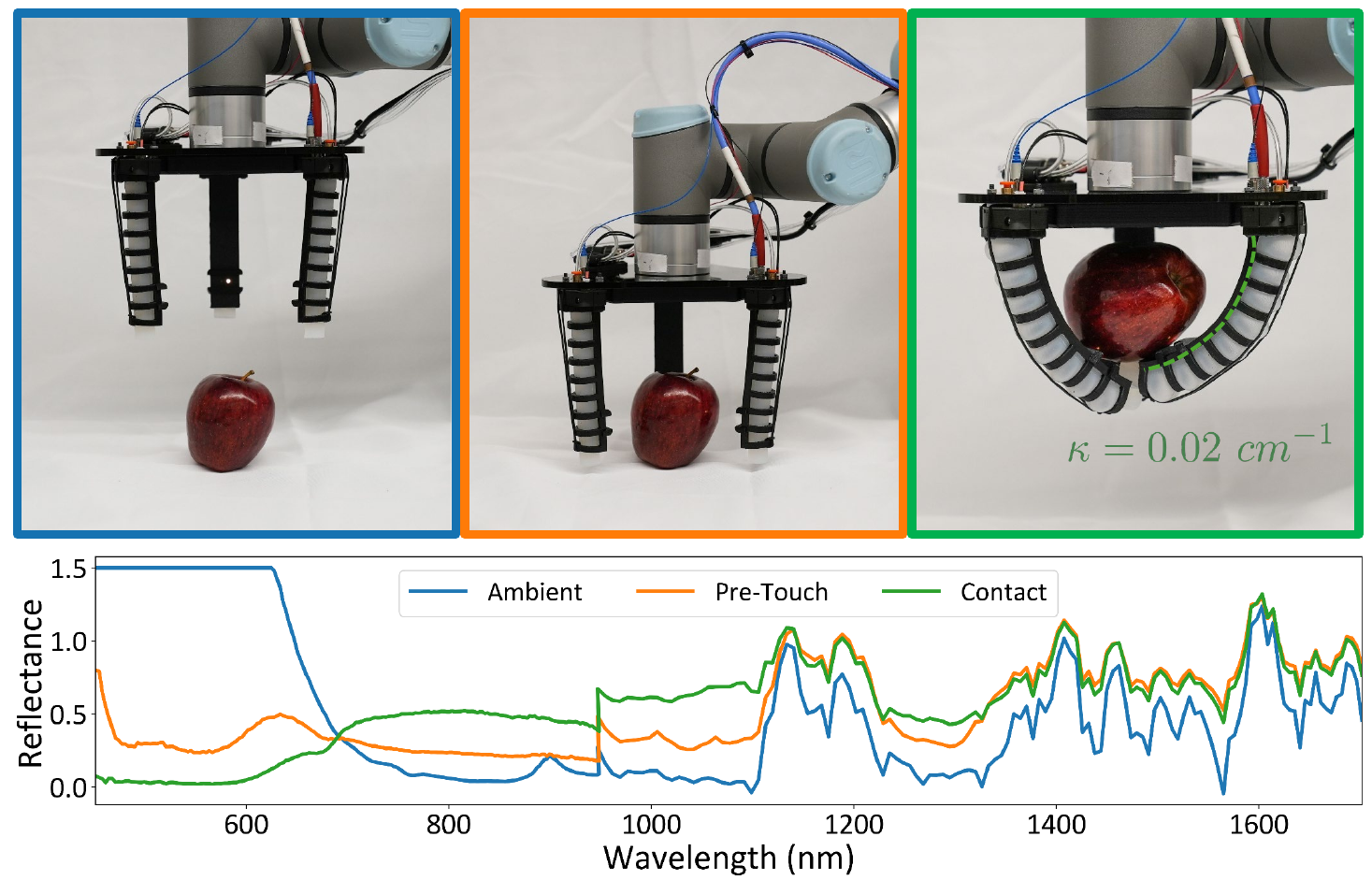}
  \caption{The top three pictures show distinct points during a grasp cycle, where the spectral signal (bottom) is acquired for the ambient environment, caging pre-grasp, and in-hand response.}
    \label{fig:scans_teaser}
    \vspace{-1.75em}
\end{figure}
\begin{figure*}[!b]
  \vspace{-1.5em}
  \centering\includegraphics[width=\linewidth]{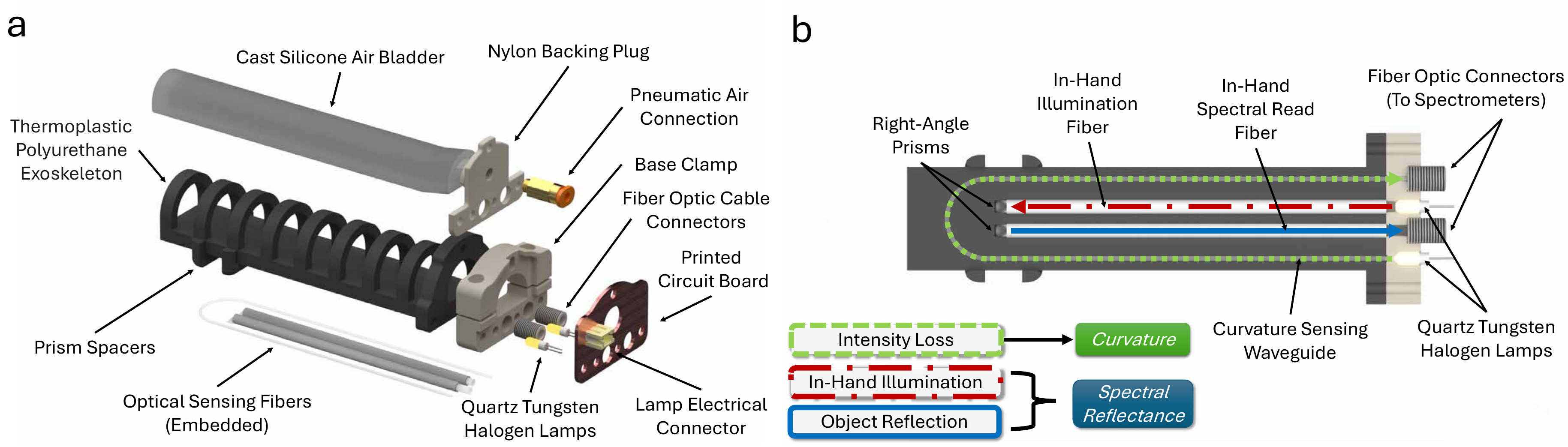}
  \caption{(a) Exploded 3D rendering of the components in a single SCANS finger. (b) Operating principle for sensor with waveguide and dual PMMA waveguides for in-hand spectroscopic measurements.}
  \label{fig:scans_cross_section}
\end{figure*}
Our present work builds upon recent advances in soft optical sensing and robotics-centric spectroscopy to introduce the soft curvature and spectroscopy (SCANS) gripper. SCANS leverages optical fibers to estimate finger curvature and perform in-hand spectral analysis without the need for embedded electronics or interconnects within the actuated portion of the finger---an area where components are often susceptible to mechanical failure. As such, the gripper acquires high-fidelity material signatures in a fully flexible and mechanically robust platform. This co-location allows for adaptive, safer interaction with fragile objects, and maintains high-fidelity spectral sensing without separate rigid components or electronic interconnects as in prior rigid systems \cite{erickson2019classification,hanson2022vast}. Figure~\ref{fig:scans_teaser} illustrates the SCANS gripper and shows the recorded spectral reflectance from an apple before contact and during grasp.

With the SCANS gripper, we address the major challenge of designing a spectral sensing finger with sufficient optical transmission across wavelengths from visible light (400 nm) to short-wave infrared (1700 nm). 
Our key contributions are:

\begin{itemize} 
    \item Infrared spectral characterization of common soft robotic materials to inform optical sensing design.
    \item A soft sensing architecture for finger curvature estimation and spectral signature acquisition.
    \item An extensible soft gripper design enabling pre-grasp object identification via integrated spectroscopy.
\end{itemize}

\section{Background and Related Work}

Multiple sensing approaches, including visual \cite{bell2015material, zheng2024materobot}, haptic interaction \cite{sinapov2011vibrotactile, kerr2018material}, and thermal sensing \cite{bai2022analyzing} enable the inference of an object's properties towards informing how it should be grasped or handled. 
While broad inter-class material classification has shown promise, visual sensors combined with neural networks often struggle to differentiate items that lack distinct textures or geometric patterns.

Approaches that require contact for material information (e.g., haptic interaction) may damage flexible or fragile objects.
Spectroscopy, in contrast, is a non-contact method that measures light intensity reflected from a surface across multiple wavelengths and can observe intra-class material composition \cite{pasquini2018near}.
Spectroscopy has been used in robotics to differentiate between generalized material classes such as metal, plastic, wood, paper, and fabric \cite{erickson2019classification} and surface terrains \cite{hanson2022vast}.
These approaches have also been integrated into manipulators \cite{hanson2022slurp, cortes2017integration, li2023underwater, hanson2022occluded} to classify minerals, object classes, and visually similar liquids and granular solids.

Within soft robotics, the bulk of optical sensing work has focused on applying soft and deformable optical waveguides to measure curvature \cite{wang2018toward, teeple2018soft}, strain \cite{to2018soft}, or surface texture \cite{zhao2016optoelectronically}.
This approach uses light intensity change at one wavelength to infer mechanical deformation. 
Because these waveguides are often designed with intentional surface imperfections \cite{teeple2018soft, to2018soft, to2015highly}, increased mechanical deformation decreases the intensity of observed light. 
The use of novel materials (e.g., hydrogel \cite{lee2020hydrogel}, thermal polyurethane \cite{han2024bio}, reflective polymers \cite{lantean2022stretchable}, self-healing polymers \cite{bai2022autonomous}) can further improve sensitivity or deformability over polydimethylsiloxane, urethane, hydrogels, silicone, resin \cite{trunin2025design}, or silica-glass fiber optic cables \cite{galloway2019fiber, to2018soft}. 
By modifying the underlying structure to achieve a new sensing modality through surface roughness \cite{wen2020encoder}, multi-color waveguides \cite{zhong2025calibration}, or gaps for materials sensing \cite{gerald2022soft}, researchers have expanded soft waveguides beyond simple deformation sensors into multiple functions. 
While these approaches illustrate the potential for applying soft optical waveguides to a range of challenges, soft optical sensing remains limited in its ability to discern the underlying material properties of a grasped object. 

Few demonstrations have focused on spectroscopy in soft robotics due to the challenge of diminished illumination on grasped surfaces, which minimizes the signal generated by observed objects.
 A prior soft gripper for spectral sensing \cite{Hanson2022-Flexible} used a flexible circuit board and halogen lamps embedded within an elastomer finger pad. Although this work approaches the problem of sensing in-hand spectral signatures, it has low optical throughput at wavelengths greater than 900 nm and cannot perceive many chemical identifiers present in the NIR range. Furthermore, embedded light sources are shown to generate a significant amount of heat ($>35^{\circ}$ C) and are mounted to an FR4 flexible PCB that is subject to fatigue failure. Finally, the design only uses light to perceive an in-hand reading; there is no ability to infer the shape of the gripper using the sensor array.

Achieving spectral analysis in highly deformable waveguides will require approaches that mitigate and compensate for light loss with deformation.
This project builds upon prior efforts to improve optical sensing in soft robotics through the selected material and signal processing approaches to ensure reliable object measurements with an expanded capacity to sense in the NIR spectrum. 
To the best of our knowledge, this research is the first to use light multi-functionally for both gripper state estimation and material identification.

\section{Design, Materials, and Fabrication}

\subsection{SCANS Finger Design and Operating Principle}

Fig.~\ref{fig:scans_cross_section} depicts the components and operating principle of the SCANS finger.
The U-shaped waveguide transmits light from a light source to a fiber optic cable. As the finger bends, light is scattered in the fiber; the intensity is used to infer the curvature of the gripper. 
For material sensing, light is transmitted from a second light source through the spectroscopic optical fiber, then reflected by the prism to transmit light perpendicular to the finger surface. 
The reflected light from an object held in hand travels through a second prism and fiber, then into a fiber optic cable to the spectrometer. To ensure enough reflected light enters the second prism, we used standoffs around the prism windows to hold the grasped object at a consistent distance.
The raw readings from the object reflection are used to infer the class or material of the grasped object. 

The finger is composed of an air bladder and a flexible exoskeleton based on prior work \cite{oliveira2019hybrid}. 
Both components have a D-shaped cross-section, which creates constant curvature bending in one axis when the bladder is inflated.
Spectroscopic waveguides---a prism and the U-shaped curvature sensing waveguide---are embedded within the exoskeleton. This dual-use architecture allows the type of optical pathway to support both curvature sensing and material identification in a soft, compliant form factor.

\subsection{Waveguide Material Selection}
Though quartz glass is most common for spectroscopy due to its transmission properties, it has an ultimate bending radius that is too large for most soft robotics applications. The selection of waveguide material will influence the ultimate bending radius, light transmission, and the ability to withstand repeated deformation. An ideal waveguide should meet the following engineering requirements: (a) consistently high transmission across the VNIR spectra, (b) high transmission above 900 nm where spectroscopic material identification occurs, and (c) a large acceptance cone to allow for signal acquisition during close-range measurements.

\begin{figure}[!tbp]
  \centering
  \vspace{0.5em}
  \includegraphics[width=\linewidth]{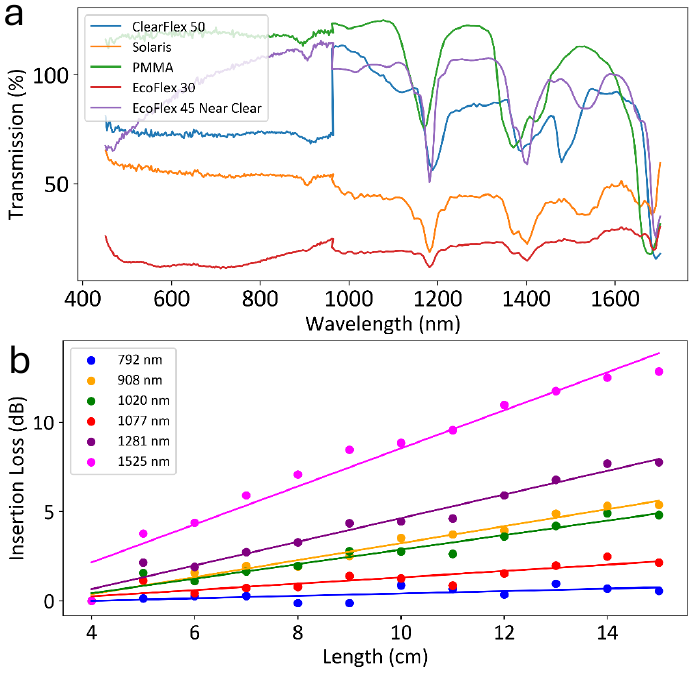}
  \caption{(a) Optical transmission profiles for each of the candidate materials for the creation of the soft sensors. These readings are converted from raw digital counts by normalizing against the transmission profile of a 6.5 mm piece of uncoated Borosilicate glass. (b) Insertion loss calculated using Eq.~\ref{eq:insertion_loss} for lengths of PMMA fibers 4-15 mm in length. Fit regression lines for each length show differing loss rates. Lower loss values are desirable, as increased loss corresponds to lower perceived signal.}
  \label{fig:scans_absorbances}
  \vspace{-1.5em}
\end{figure}

\paragraph{Transmission}
To find a suitable material that is mechanically robust and optically suitable, we evaluated the optical transparency of EcoFlex 30, EcoFlex 45, Solaris, Clear Flex-50, and polymethyl methacrylate (PMMA).
First, we directed a fiber-coupled quartz tungsten halogen (QTH) light source through a 6.5 mm thick sample of each material. 
The readings are min-max normalized against the transmission through an uncoated, 6.5 mm thick borosilicate glass disc that approximates the transmission expected through a quartz fiber optic cable (Fig.~\ref{fig:scans_absorbances}a).

EcoFlex 30, a translucent material, appreciably scatters light in the visible range but approximately matches the transmission profile of other silicone-based elastomers (EcoFlex 45 Near Clear and Solaris) in the NIR. Clear Flex has excellent optical throughput with one additional absorption trough at 1475 nm. The PMMA fiber performs the best in the VNIR (400--1000 nm) range but has larger absorption bands at 1175, 1400, and 1650 nm. 

\paragraph{Insertion Loss}
To ensure transmission of sufficient light through the waveguide as the waveguide length is increased, we quantify the insertion loss: the light lost per unit length of a waveguide. As it relates to transmission, insertion loss quantifies the scalability of the waveguide material to different finger sizes and how perceived spectral intensity will vary as a function of the transmitted light path in-hand. We observed the insertion loss by measuring the spectral intensity through PMMA fibers coupled to a QTH light source and a visible-to-short wave infrared (VIS-SWIR) spectrometer. We plot loss as a function of length in Fig.~\ref{fig:scans_absorbances}b using an insertion loss calculation as

\vspace{-1.5em}
\begin{gather}
    \label{eq:insertion_loss}
    \text{Loss (dB)} = - 10 \log_{10} \frac{S_i(\lambda)}{S_0(\lambda)}
\end{gather}

where $S_i(\lambda)$ represents the digital counts for fiber of length $i$, and $S_0(\lambda)$ approximates transmission through 4 cm of fiber. The loss increases linearly with length at all wavelengths, but the rate of loss is wavelength-dependent. For example, light at 792 nm is lost at 0.07 dBcm$^{-1}$ while at 1525 nm it is lost at 1.07 dBcm$^{-1}$. Comparing these PMMA absorptions with the other materials in Fig.~\ref{fig:scans_absorbances}a, we note areas of strong absorbance, as these wavelengths will correspond to low signal-to-noise ratios in longer waveguides. The higher the initial transmission profile, the longer the waveguide will need to be before the loss becomes significant.

\paragraph{Numerical Aperture}
A large numerical aperture (NA) improves light transmission and reduces losses from misalignment. 
Complementary fiber core and cladding refractive indices permit total internal reflection and light transmission through the waveguide. 
The critical angle \( \theta_c = \arcsin{(n_{\text{clad}}/n_{\text{core}})} \) defines the minimum angle for total reflection and therefore the maximum bend radius of the finger before all light transmission is lost.
Finally, the acceptance cone, defined by the NA, determines the range of angles of incident light accepted by the fiber and is influenced by the refractive indices $n_{\text{core}}^2$ and $n_{\text{clad}}^2$, of the core and cladding:
\[
NA = \sqrt{n_{\text{core}}^2 - n_{\text{clad}}^2}
\]
PMMA fibers minimize light loss from radial compression, making them suitable for SCANS sensors. For optimal performance, \( n_{\text{clad}} < n_{\text{core}} \) to enhance transmission and minimize cladding losses. These fibers have a PMMA core with a refractive index of 1.50 and an outer cladding of fluorinated polymers with a refractive index of 1.40. The NA of these fibers is 0.54 and the critical angle, $\theta_c$, is 69$^{\circ}$.

\subsection{Fabrication}

\begin{figure*}[!t]
  \centering
  \vspace{0.5em}
  \includegraphics[width=\linewidth]{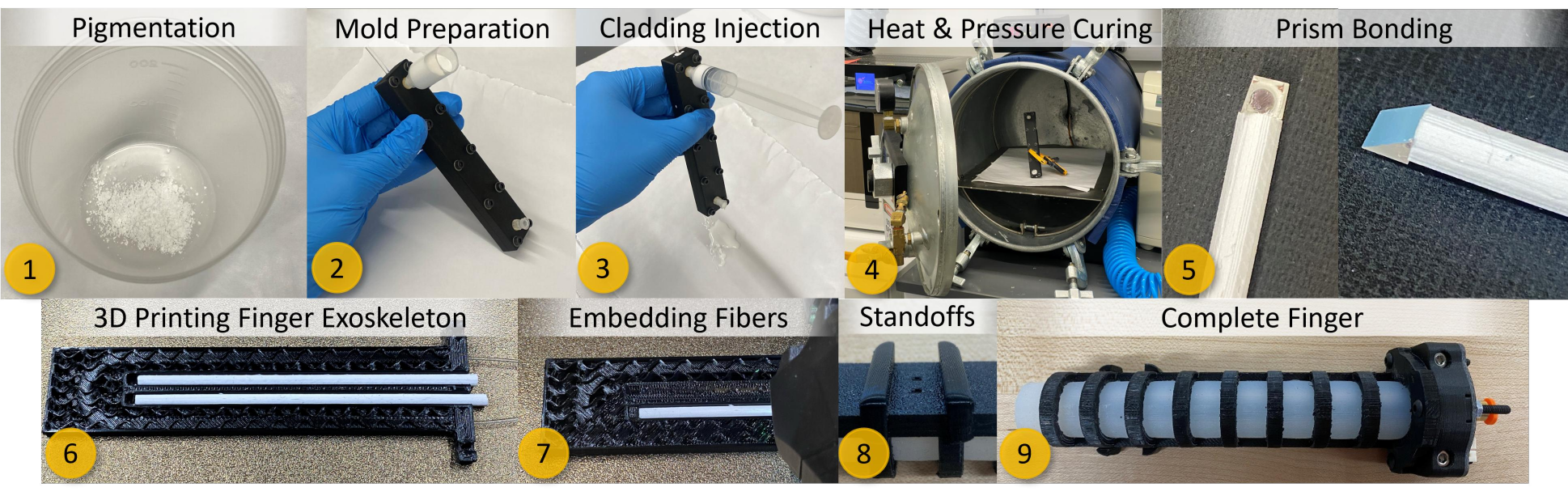}
  \caption{Fabrication process for single SCANS waveguide from raw materials to integration in a soft robot finger.}
  \label{fig:scans_fabrication}
  \vspace{-2.00em}
\end{figure*}

Fig. \ref{fig:scans_fabrication} illustrates the process of preparing the spectroscopic waveguide cladding and affixing it to the prisms (steps 1--5).
The spectroscopic waveguides are comprised of a 2.5 mm diameter PMMA fiber (Azimom), a 0.5 mm thick injection-molded cladding, and a 3.0 mm micro-machined prism (Tower Optical) with an aluminum-coated hypotenuse. 
The cladding is DragonSkin\textsuperscript{TM} 10 NV (Smooth-On), which is thinned with 7.5\% of the total mass of silicone thinner (Smooth-On). Based on the results from \cite{lantean2022stretchable}, we also add titanium dioxide ($\text{TiO}_\text{2}$) pigmentation to act as a secondary cladding. The mass of $\text{TiO}_\text{2}$ powder is added at 5\% of the mass of the DragonSkin\textsuperscript{TM}. This mixture is combined using a planetary centrifugal mixer (Thinky USA) for 2 minutes to evenly disperse the pigmentation and to remove air bubbles. The mixture is then injected into the mold using a syringe and placed inside a pressure molding chamber (X-11, Hapco) at 550 kPa and 60 \degree C for one hour. The mold clamps the fiber at both ends and applies a constant tension during curing.

The fiber is then manually demolded, deflashed, and cut to size with a razor blade and a guide to ensure a 90-degree cut. After two fibers are prepared, they are glued to the prism's side using a 3D printed alignment jig. A small bead of epoxy adhesive DP100 (3M) is dispensed on the prism face using a 1:1 50 mL dispensing gun, and the combined part is set for 20 minutes. The U-shaped curvature waveguide is a 1.5 mm diameter PMMA fiber (Azimom) without cladding.

Next, we 3D print the flexible exoskeleton using a 95A shore hardness TPU (ZIRO) with a 0.4 mm hardened tool steel nozzle at a 0.1 mm layer height. The curvature waveguide and the spectroscopic fiber-prism assemblies are embedded in two different layers. To embed the optical sensing fibers, we pause the print at the layer just above the tallest point of the component. The fibers are laid inside the TPU exoskeleton, and the print is resumed (Fig. \ref{fig:scans_fabrication}, steps 6-7). Once the exoskeleton print is complete, support material is removed, and the waveguide fibers are trimmed using a 3D printed jig and a razor blade.
\begin{figure}[!b]
  \centering
  \includegraphics[width=\linewidth]{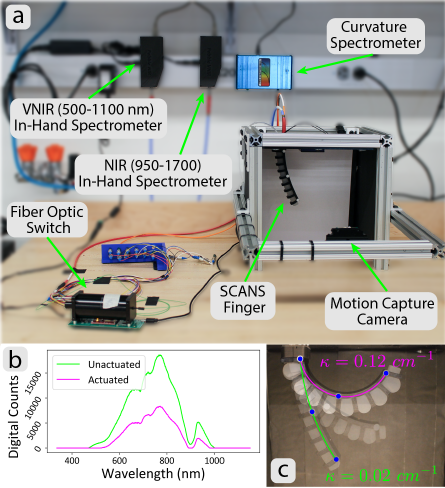}
  \caption{(a) System architecture used to quantitatively evaluate the SCANS sensor performance. (b) Associated curvature waveguide spectral response in unactuated and actuated states as shown in (c). Note the decrease in spectral intensity. (c) Motion tracking profile of the finger in unactuated and actuated states with the associated curvature measurements in radians.}
  \label{fig:scans_testbed}
\end{figure}

\begin{figure*}[!t]
  \centering
  \vspace{0.5em}
  \includegraphics[width=\linewidth]{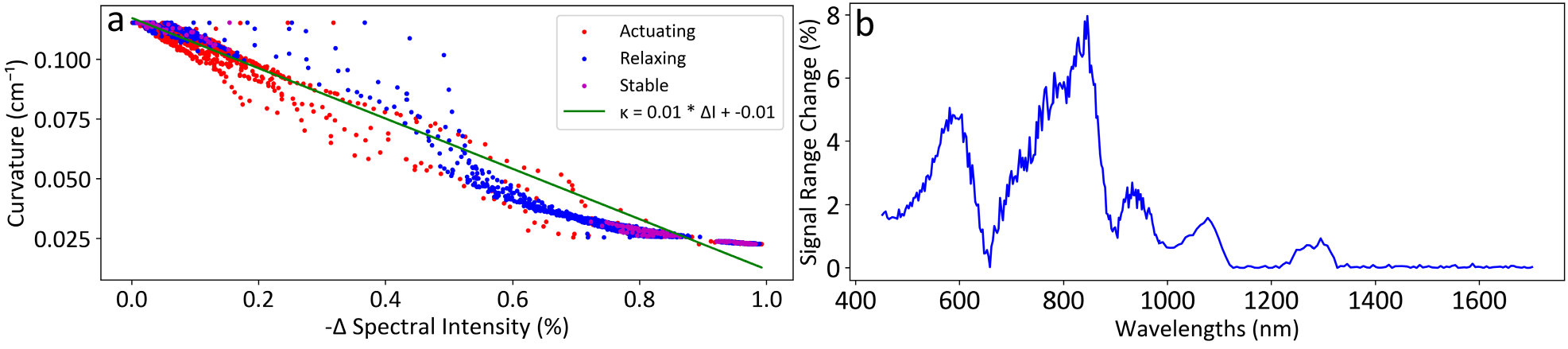}
  \caption{(a) Regression model with the predicted curvature as a function of the observed change in optical intensity. (b) Modeling of signal attenuation for in-hand spectral sensing capabilities after 100 full actuation grasping cycles. }
  \label{fig:curvature_regression}
  \vspace{-1.5em}
\end{figure*}

The optical path of light from the illumination to the read fiber requires a 5 mm gap between the prism and the object. Without this gap, no light from the illumination fiber would reach the read fiber. To prevent this contact, we use two thin prism spacers (Figure \ref{fig:scans_absorbances}a) secured with Loctite 495\textsuperscript{\textregistered} (Figure \ref{fig:scans_fabrication}, step 8). This air gap allows the illuminated surface to align with the read fiber's optical acceptance cone. The spacer height was chosen to achieve a 50\% surface area overlap at contact.

Connections between the air bladder, waveguides, exoskeleton, and offboard power and fluidic sources are made with a clamp at the end of the finger printed in carbon fiber nylon (Onyx\textsuperscript{\textregistered}, Markforged). The silicone air bladder is manufactured according to \cite{oliveira2019hybrid} and trimmed to size using a razor and two 3D printed jigs. It is then affixed to the 3D printed clamp with a thin coating of Sil-Poxy\textsuperscript{TM} (Smooth-On) on the nylon backing plug in Fig. \ref{fig:scans_cross_section}a. One end of the curvature fiber and one of the prism-bonded fibers are mounted flush with SMA adapters (Thorlabs) which are used to couple the PMMA fibers with low-OH (i.e., hydroxyl), 1000 $\mu$m, 0.5 NA glass fiber optic cables (Thorlabs). These cables then connect the embedded waveguides to spectrometers while minimizing the amount of light lost in transit. The large NA of these fibers increases tolerance for fiber-waveguide misalignment. The bladder is then inserted into the exoskeleton, and the exposed fibers are inserted into the clamp. Finally, the PCB housing the QTH bulbs (1150-9A, International Light Technologies) is affixed to the clamp. The bulbs are mounted on the PCB and sit flush to the input fibers.

\section{SCANS Sensor Characterization}
With a full SCANS sensor, we now turn to characterizing the performance of the system under actuation. 
We designed experiments to uncover the relationship between optical signal intensity across wavelengths with finger curvature and to understand which regions of the spectrum permit differentiation between objects.

\subsection{Measurement Architecture}

Fig.~\ref{fig:scans_testbed}a shows the system architecture used to characterize and conduct spectroscopic measurements through the SCANS sensor. The output fiber optic cable for the in-hand sensing is coupled to a bifurcated fiber optic cable (Ocean Insight) which directs the signal from the output into a visible-near infrared (VNIR) spectrometer (Pebble VIS-NIR, Ibsen Photonics) and a NIR spectrometer (Pebble NIR, Ibsen Photonics). This spectrometer pairing is further detailed in \cite{hanson2023hyperdrive}. The curvature sensing fiber is coupled into a separate spectrometer (Blue Wave VIS, StellarNet). When multiple fingers are operated together, such as in the gripper configuration of Fig.~\ref{fig:scans_teaser}, a fiber optic switch (Fiber-Fiber 1x16 Mini Switch, Agiltron) cycles the optical path between each of the waveguides and the spectrometer. Regulated air pressure at 200 kPa is supplied to a fluidic control board (Soft Robotics Toolkit) to control the actuation of the finger. The integration time for the curvature spectrometer is set to 1 ms to avoid photodetector saturation. All device drivers and the actuation of fluidic control board solenoids are managed by an Ubuntu 20 PC with ROS \cite{quigley2009ros}. 

\subsection{SCANS finger performance}
\paragraph{Curvature Characterization}
The procedure of curvature estimation via measuring optical intensity changes as the finger bends is well established in the literature \cite{aljaber2024optimal,teeple2018soft}. 
To estimate curvature, we assume the finger will bend only around a primary axis and that it will conform to a circular arc with radius $R$ and length $l$. Mirroring the literature \cite{aljaber2024optimal}, we report curvature as $\kappa = \frac{1}{R}$.

To study this model fit, a SCANS sensor was instrumented with three reflective marker dots at the tip, midpoint, and base of the finger along the constant length $l$. A motion capture camera \cite{lvov2023mobile} tracked the position of these markers in real-time as the finger was actuated and curved. Fig.~\ref{fig:scans_testbed}b shows the marker positions during actuation. Fig.~\ref{fig:scans_testbed}c shows the waveguide spectra at maximum and minimum curvature. Ground truth measurements were published at 30 Hz, and the spectrometer was sampled at 100 Hz.

Although the curvature waveguide intensity is read out with a spectrometer, the intensity at one wavelength is sufficient to model curvature. We chose 875 nm because it exhibited a strong signal (low absorption) when the finger is unactuated (Fig.~\ref{fig:scans_absorbances}). 
Fig.~\ref{fig:curvature_regression}a is a plot of the curvature estimated from the optical ground truth against the intensity normalized to the  unactuated intensity. We fit a linear regression model to predict curvature from intensity change and achieve an $R^2$ value of 0.98. 
Notably, the curvature-intensity linear fit has the same trend for both increases and decreases in curvature.

\paragraph{Fatigue-Induced Loss}
We evaluated the sensor's resilience to repeated actuation cycles by investigating the decay in intensity over repeated actuation cycles. Starting with a new sensor, we measured the light reflected from a Spectralon reference standard. We then actuated the finger for 100 cycles and re-measured on the standard. Fig.~\ref{fig:curvature_regression}b shows the results of this modeling. The finger experiences less than a 10\% loss in sensing range at all wavelengths. Most critically, the observed spectra lose less than 2\% intensity from 1000--1700 nm. This region contains most of the known chemical absorbances that are critical to differentiate material \cite{pasquini2018near}. Externally, there is no visible delamination between the fibers and the prisms, which would appear as light leaking through the gripper exoskeleton.

\label{sec:experiments}
\begin{figure*}[!t]
  \centering
  \vspace{0.5em}
  \includegraphics[width=\linewidth]{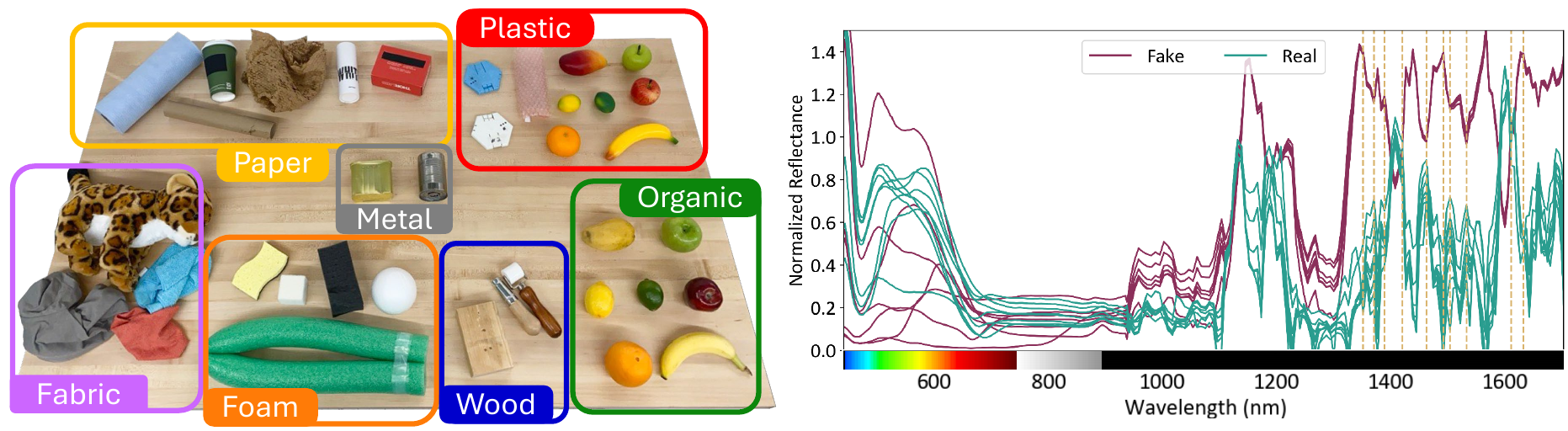}
  \caption{(Left) Items used in grasping trials with categorical labels. (Right) Plots of mean spectra for each of the faux and real fruits. The gold bars on the spectral plot correspond to the wavelengths that most significantly contribute to in-hand differentiation in a binary LDA.}
  \label{fig:scans_experiments}
  \vspace{-1.5em}
\end{figure*}

\paragraph{Pre-Grasp Sensory Perception}
Using the test stand from the prior section, we quantify the sensor's ability to perceive items without direct contact, as depicted in Fig.~\ref{fig:scans_teaser}. We 3D print PLA spherical sections with diameters of 95, 115, 135, and 155 mm and mount them adjacent to the finger's base. This positioning emulates the gripper enveloping an item before grasping. The gripper then performs 10 grasp cycles around the sphere. During each of these grasps, we record the final spectral measurement under contact as the desired signal. As the finger is actuated to grasp, we observe the correspondence of the contact measurement with the finger in various stages of actuation---unactuated to final contact.

We use the spectral angle mapper (SAM) \cite{kruse1993spectral} to assess the correspondences between spectral measurements. SAM evaluates the similarities in the shapes of spectra, focusing on the relative spacing of peaks and troughs corresponding to spectral absorbances, rather than the average signal intensity.  In (\ref{eq:sam}), $r$ is a reference spectral measurement to compare to, and $c$ is the current measurement.

\begin{equation}
    \label{eq:sam}
    SAM(r,c) = \arccos{\left( \frac{
     r\cdot c
    }{ ||r|| * ||c||  }\right)}
 \end{equation}

For each size object, we take the final in-hand spectral measurement once the finger has stopped moving and compare it to the in-hand spectral measurements for when the gripper is at 0\%, 25\%, 50\%, 75\%, and 90\% of its final curvature using SAM. We report the mean and standard deviations of the SAM scores across all data samples, as well as the average calibrated signal intensity in Table~\ref{tab:pregrasp_measurements}.

\begin{table}[b]
    \centering
    \resizebox{\columnwidth}{!}{
        \begin{tabular}{lcccccc}
            \toprule
            \textbf{Metric} & \textbf{No Curve} & \textbf{25\% Curve} & \textbf{50\% Curve} & \textbf{75\% Curve} & \textbf{90\% Curve} \\
            \midrule
            $\overline{\text{SAM}}$ & 12.60 & 12.43 & 8.15 & 6.43 & 4.34 \\
            $\sigma(\text{SAM})$    & 1.13  & 1.39  & 3.31 & 2.10 & 1.86 \\
            $\overline{S_{\text{cal}}}$ & 0.16  & 0.19  & 0.20 & 0.20 & 0.20 \\
            \bottomrule
        \end{tabular}
    }
    \caption{\textbf{Pre-grasp measurement consistency} calculated using spectral angle mapper between final in-hand, contact spectral measurement for all objects. All measurements reported in degrees.}
    \label{tab:pregrasp_measurements}
\end{table}

Similar to the conclusions from \cite{hanson2024prospect}, the distance and glancing angle of an optical aperture affect the consistency and amplitude of the observed spectral measurement. Ideally, as the gripper closes, the acceptance cone of the fibers will move towards the object, and proportionally less light from the ambient environment will be coupled into the optical fibers. Fig.~\ref{fig:scans_teaser} graphically depicts this evolution of the spectra. Initially, the pixels in the visible range saturate as ambient environmental illumination adds to the signal. As the finger curves toward the object, we observe a corresponding decrease in the mean of the observed SAM. This increasing consistency in measurements corresponds to a decreasing spot size of the projected measurement. For objects with heterogeneous material composition, this is a notable consideration as multiple surface points may be scanned depending on the geometry of the object. Finally, we note the increase in $\overline{S_{cal}}$ as the finger nears its final grasping state. In an unactuated state, the finger is susceptible to noise from lights other than the in-hand illumination; however, indoor lights tend to be biased toward the visible spectra, which results in a lower intensity NIR spectra. The increase in intensity mirrors the lighting distribution normalizing as a higher proportion of the finger's active illumination becomes incident on the object.

\section{Grasping Experiments}
The sensorized finger is a standalone unit that can be configured to form a variety of parallel and anthropomorphic grippers. Three of these soft fingers are mounted in a triangular pattern to form a gripper (Fig.~\ref{fig:scans_teaser}). The three waveguide outputs are connected to the fiber optic switch, and inputs are cycled at a rate of 10 Hz. This allows a single spectrometer to observe the curvature of all three fingers. In the actuated state, the three finger tips overlap each other to form a caging grasp. We ensure consistent spectral measurements by actuating a single finger 3 seconds before the other two. This grasp pre-forming allows one finger to make contact with the object before the other two come together to envelop it in a stable grasp. As this current scope of work assumes each item will constitute a uniform material, we only record the measurement of the first spectral finger to contact the object. We add a set of elastic bands to the gripper exoskeleton to act as a restoring force to the fingers and maintain a wide aperture.

\paragraph{Inter- and Intra-class Object Differentiation}
We validate the spectral sensitivity of the SCANS gripper by showing the system's ability to provide inter-class differentiation between objects of similar material classes. Our dataset includes grasps of the following materials with varying colors (Fig.~\ref{fig:scans_experiments}): \textit{Plastic, Metal, Fabric, Wood, Foam, Paper}, and \textit{Organics} totaling 37 items. This set of items constitutes items commonly found in domestic settings. Additionally, we add visually confusing items, including real and fake fruits, to demonstrate the gripper's ability to distinguish non-visual characteristics.

For each object, we collected three samples, rotating the objects relative to the gripper to ensure variation in contact points. Measurements are collected from the time of the initial finger movement to the final caging. As the in-hand spectrometers report measurements in raw digital counts, we perform a reflectance calibration routine (\ref{eq:reflectance}). 
\begin{equation}
    \label{eq:reflectance}
    s_{\text{cal}} =\frac{s_{raw} - s_{dark}}{s_{white} - s_{dark}}
\end{equation}

First, a piece of calibrated white diffuse reference material (Spectralon 99\%) is placed 20 mm in front of an unactuated finger to generate the maximum possible intensity signal, $s_{white}$. We adjust the spectrometer integration times to avoid saturation under the white reference signal. The VIS-NIR and NIR spectrometer integration times are 25 ms and 250 ms, respectively. Next, we turn the system lamps off and cover the return optical path to yield the dark reference signal $s_{dark}$. These two values are saved, and each subsequent measurement, $s_{raw}$, is normalized against them to generate $s_{cal}$. This routine is performed at the start of all grasping experiments, with consistent ambient light. Although ambient light can affect the spectral reading, the measurement is primarily governed by light reflected from the illumination fiber during pre-touch and contact.

 We calculate the pair-wise SAM to understand the similarity of the perceived signals. Fig.~\ref{fig:scans_sam} shows the similarity between every item combination grouped together and averaged as a function of the larger material class from Fig.~\ref{fig:scans_experiments}.
 
\section{Discussion}

Fig.~\ref{fig:scans_experiments} shows the average spectral reflectance-normalized profiles extracted for the real and fake fruits. In comparison to prior work \cite{Hanson2022-Flexible}, this system overcomes the strong absorption and noise present at wavelengths greater than 900 nm. This added sensing range enables the detection of additional chemical absorbances in the NIR \cite{pasquini2018near}.

The real fruit spectra in Fig.~\ref{fig:scans_experiments} are similar to the faux fruit under visible light. Independent of the color under visible light, the real fruit also exhibits strong separation of the spectral signatures in the NIR with particularly distinct absorbances prevalent at 900, 1150, 1370, and 1630 nm. To identify which wavelengths contribute most to distinguishing real from faux fruits, we apply linear discriminant analysis (LDA) as an interpretability tool. The discriminant coefficients highlight the spectral regions most relevant for class separation, following standard practice to interpret feature importance, even when classification is not the primary goal \cite{ye2004two}.

After constraining the dimensionality reduction of LDA to one component, the 10 largest coefficients, which correspond to the importance of the wavelength in the decision function, are plotted in Fig.~\ref{fig:scans_experiments}. The most informative wavelengths all lie within the NIR spectra. This result shows the importance that chemical absorbances play in distinguishing between items of differing chemical composition; moreover, it validates the ability of the SCANS gripper to obtain meaningful spectral information for future machine learning models. While our experiments use uniform materials, real-world objects often contain mixed surface and subsurface properties (e.g., coatings or interior fills). Prior work shows that machine learning can help interpret such complex spectra \cite{hanson2022slurp}.

\begin{figure}[!t]
  \centering
  \vspace{0.5em}
  \includegraphics[width=\linewidth]{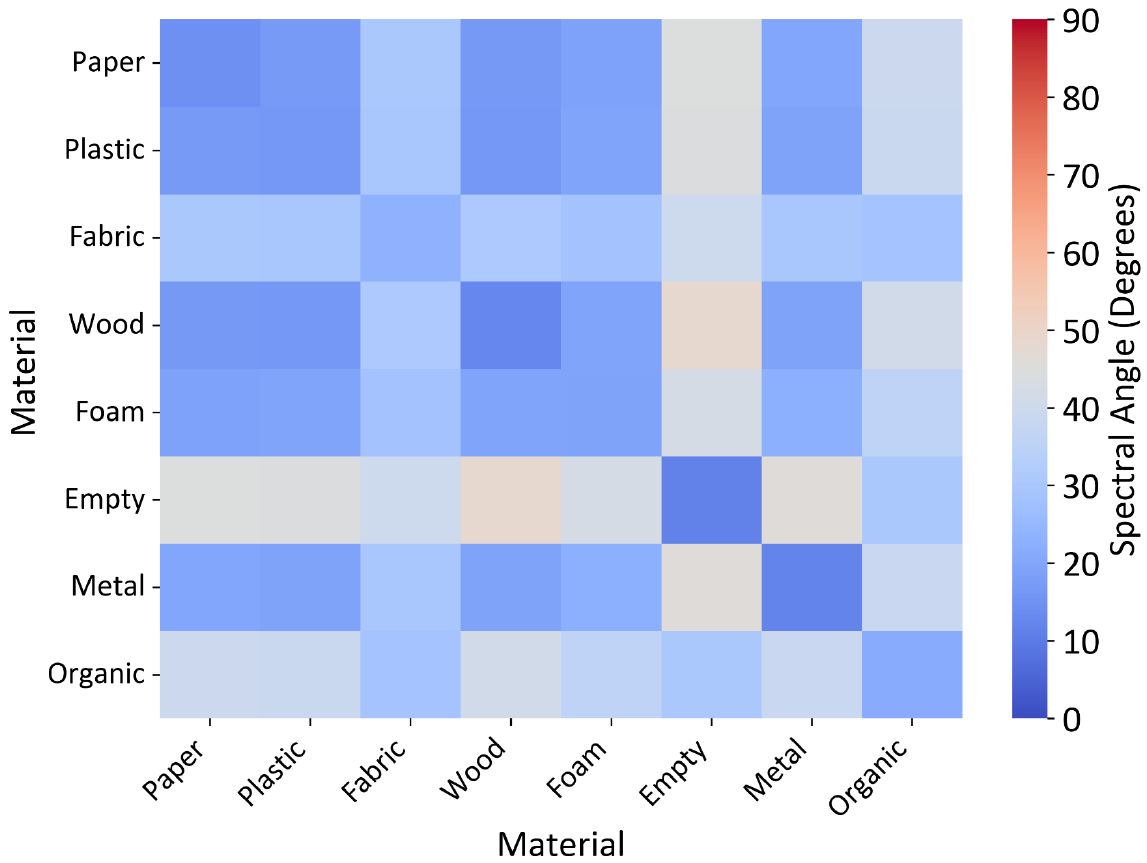}
  \caption{Spectra angle mapper (SAM) calculated between all collected spectral signatures grouped by material class. A larger score indicates there is larger distinction between the shapes of the spectral signatures.}
  \label{fig:scans_sam}
  \vspace{-1.5em}
\end{figure}

 One notable observation in Fig.~\ref{fig:scans_sam} is the strong change that occurs when an object is brought in front of the sensing aperture. The row labeled \textit{Empty} indicates the spectral signature of an ambient measurement. This is an appreciable response and allows the distinction between when the optical aperture is open, as in the case of any empty grasp, or when there is an object present in hand. The \textit{Empty} grasp has a high degree of variability with every other class of spectral measurement. This observation is useful for confirming whether a grasp has been successfully completed and an object is held within the fingers of a gripper.

As seen in Fig.~\ref{fig:scans_experiments}, the organic items exhibit a large difference from every other class. These results consider the application of $SAM$ to the whole spectra, including the object color. This difference is due to the presence of water, which has several strong spectral absorbances in the NIR. In general, a SAM score of $\approx20\degree$ or greater indicates sufficient inter-class differences to separate the two measurements using a machine learning model \cite{petropoulos2010comparison}. While machine learning exploitation is not considered in the scope of this paper, our results show the in-hand spectral data have great potential for deep learning models, as simple statistical measures can derive inter-class differences not readily apparent through color vision.

Our curvature model assumes in-plane, inward bending and may be less accurate with large or irregular objects. For highly compliant materials such as foam or organics, surface deformation around the prism spacers may influence spectral measurements. These interactions present opportunities to further explore how deformable contact conditions affect sensing performance.

Curvature measurement via a fiber optic switch functions well; however, the spectral intensity drops when the internal switch mirror repositions to redirect the correct input in-hand fiber to the spectrometers. Because we identified wavelengths with high curvature sensitivity through our experiments, a future version of the system could use narrow-bandwidth photodetectors at these target wavelengths and eliminate complexity in the measurement architecture.

\section{Conclusions}
We introduced a first-of-its-kind spectral sensor that uses optical components to infer grasp state and spectral readings of grasped items. We provided a detailed analysis of the properties necessary for high-throughput optical fibers and their suitability for inclusion in soft robotics. We also detailed a novel fabrication procedure to construct the highly compliant, electronics-free sensor with silicone casting and additive manufacturing. We showed the system’s capabilities with a cyclic test of the curvature sensing waveguides to assess short-term sensor performance and stability, and the differentiation of items using the in-hand spectral signatures.  Further extended testing will be valuable for evaluating long-term durability in higher-cycle, industrial use cases.

The PMMA fibers in the SCANS gripper demonstrated excellent light transmission. Additional material engineering could eliminate some of the strong absorbances present in the NIR spectrum. While this work largely focused on exploiting the entire spectra to differentiate between grasped objects, future efforts will 1) investigate the fusion of curvature and spectroscopic measurements for improved object identification, and 2) explore targeted applications of the spectra to perform quality control in automation and identify intra-class properties, such as produce ripeness. The addition of deep learning models to process the data jointly could be used in applications such as automated fruit grading. Altogether, our work demonstrates optical sensing is a powerful and multi-functional capability for soft robots to gain a greater understanding of the objects they manipulate.



\bibliographystyle{IEEEtran} 
\bibliography{references}

\addtolength{\textheight}{-12cm}   

\end{document}